# Transformer-based Text Classification on Unified Bangla Multi-class Emotion Corpus

Md Sakib Ullah Sourav[1], Huidong Wang[1],
Mohammad Sultan Mahmud[2*], Hua Zheng[2]

[1]School of Management Science and Engineering, Shandong University of Finance and Economics, Jinan, China.
[2*]College of Computer Science and Software Engineering, Shenzhen University,
Shenzhen, 518060, China.

*Corresponding author(s). E-mail(s): sultan@szu.edu.cn;
Contributing authors: sakibsourav@outlook.com;
huidong.wang@ia.ac.cn;
zhenghua2017@email.szu.edu.cn;

**Abstract**

Due to its importance in studying people's thoughts on various Web 2.0 services, emotion classification is a critical undertaking. Most existing research is focused on the English lan- guage, with little work on low-resource languages. Though sentiment analysis, particularly emotion classification in English, has received increasing attention in recent years, little study has been done in the context of Bangla, one of the world's most widely spoken languages. In this research, we propose a complete set of approaches for identifying and extracting emo- tions from Bangla texts. We provide a Bangla emotion classifier for six classes, i.e., anger, disgust, fear, joy, sadness, and surprise, from Bangla words using transformer-based models, which exhibit phenomenal results in recent days, especially for high-resource languages. The Unified Bangla Multi-class Emotion Corpus (UBMEC) is used to assess the performance of our models. UBMEC is created by combining two previously released manually labelled datasets of Bangla comments on six emotion classes with fresh manually labelled Bangla comments created by us. The corpus dataset and code we used in this work are publicly available.

# 1 Introduction

While enough research has been done to identify emotions from visual and auditory data, emo- tion recognition from textual data is still a new and active study topic [4]. WeChat, Twitter, YouTube, Instagram, and Facebook, as well as other Web 2.0 platforms or social networks (SNs), have recently emerged as the most important platforms for social communication [32], education [23], information exchange [31], and other purposes [2, 9, 10] among a variety of people. Users of SN connect, share their thoughts, feelings, and ideas, and participate in dis- cussion groups. Text conversation, or more specifically, emotion classification (EC), is essential to comprehending people's activities since the internet's invisible nature has made it possible for a single user to engage in violent SN speech data [19].

EC is a subset of sentiment analysis (SA). Text-based SA is usually classified into two types: opinion-based and emotion-based. Text polarity, which divides text or sentences into positive, negative, or neutral feelings, is used to classify opinions [5]. EC is a technique for extracting fine-grained emotions from speech, voice, picture, or text data [8]. Understanding the emotion or sentiment behind a particular activity or trend in online content is of significant value to busi- nesses, consumers, corporate leaders, governments, and other interested parties [20] because an increasing number of people on virtual platforms are producing online material at a rapid rate. In many human-computer interaction (HCI) systems where text is the major form of com- munication, text classification is also crucial. The significant rise in SNs has caused EC to divert its attention to social media data analysis. Nowadays, it is common practice to employ com- putational linguistics, machine learning (ML), and deep learning (DL) to assess the emotions or experiences indicated in user-written comments [27]. One of the most difficult issues in natural language processing (NLP), a branch of artificial intelligence (AI) that requires a comprehension of natural language for many HCI applications [24], is classifying emotions in text.

About 228 million people speak Bangla as their mother tongue, and another 37 million do so as a second language, making it the fifth most widely used native language in the world [1].

---

[1] https://www.ethnologue.com/language/ben/

Bangla data storage has increased dramatically online recently due to the rise of Web 2.0 apps and related services, similar to other major languages. Unstructured textual formats such as re- views, opinions, suggestions, ratings, comments, and feedback are just a few examples of how this data is typically presented. Due to the supervised nature of classification approaches, more labelled data must be used for ML and DL model training to yield useful results. High-quality Bangla-labeled data is nevertheless scarce in many sectors. Contrary to English and other West- ern dialects, which are acknowledged as being rich dialects in terms of linguistics and technology, analyzing these massive volumes of data using NLP to identify underlying sentiments or emo- tions is a challenging research topic for resource-constrained languages like Bangla [12, 16]. The highly inflected elements of the Indo-Aryan language, such as its 36 different noun forms, 24 different pronoun forms, and more than 160 varied verb forms, make the EC operation in Bangla exceptionally difficult.

BNEmo [28] and BEmoC [15] are two Bangla emotion corpora that include 6327 and 7000 tagged reviews, respectively, and are divided into six groups, i.e., anger, disgust, fear, joy, sor- row, and surprise. However, the development of an automated emotion classifier for Bangla literature requires a comparatively larger amount of emotion corpus. As a consequence, the transformer model [3] is the basis of our proposed model's inspiration as we strive to con- duct the EC of Bangla phrases in this study. The aforementioned paper presented a thorough analysis of transformer-based models for emotion detection in texts using pre-trained Bidi- rectional Encoder Representations from Transformers (BERT) word embeddings. Recently, a range of downstream NLP applications have shown the efficacy of transformer-based deep neu- ral network-based architectural models and modifications, especially for resource-rich languages (e.g., English). Because Bangla EC has been the subject of some prior studies, we want to assess it in the most efficient and trustworthy manner possible.

In order to respond to the following two questions, this research study's goal is: (1) Is it possible to determine the emotion expressed by a social network user in Bangla using a transformer-based DL model? (2) Does the DL approach with BERT word embeddings work better than the ML-based approaches to EC for the Bangla language? The use of a pre-trained word embedding model for EC of Bangla texts is examined to solve the first study question. Unlike ML-based techniques, Multilingual BERT is a DL model based on pre-trained word em- bedding that is used to examine semantic links between words. The DL models were compared to Bangla EC's ML-based approaches to answer the second question.

The main contributions of our research are as follows:

• A new, larger, unified 6-class (anger, disgust, fear, joy, sadness, and surprise) EC dataset named Unified Bangla Multi-class Emotion Corpus (UBMEC) has been constructed for Bangla based on user reviews. It amalgamates two previously published publicly available Bangla emotion corpus, BNEmo and BEmoC, as well as additional manually tagged corpus by us, resulting in an adequately developed Bangla emotion corpus. It is gathered from various domains such as food, software, entertainment, politics, sports, and others.

• A multilingual BERT model (m-BERT) for Bangla EC is being fine-tuned. This model is based on a BERT base with 12 layers, 768 hidden heads, and 110 M parameters and has been trained on 104 languages, including Bangla.

• A set of baseline results from ML models, i.e., LR, NB, and XGBoost, to create a benchmark for multi-class EC in Bangla.

The remainder of the paper is laid out as follows: The related work of EC is discussed in Section 2. The recommended technique is presented in Section 3. Section 4 examines the experimental findings and assessment criteria. The paper concludes in Section 5.

## 2 Related Works

### 2.1 Transformer models in text-based emotion classification (EC)

Researchers put their efforts into building state-of-the-art transformer models that detect emotions from a text in rich resource languages, mostly English. Huang et al. [35] achieved F1 scores of 81.5 and 88.5 for the "Friends" and "Emotion-Push" datasets, respectively, while ex- amining BERT's emotion identification abilities. There are four emotion classifications in both datasets, "friends" and "emotion-push," developed in [11]. Huang et al. [35] altered BERT and created two models, FriendsBERT and Chat-BERT, for the Friends and Emotion-Push datasets, respectively, using various pre-training tasks before the official training procedure. Using a modified BERT model, Huang et al. [14] achieved 0.7709 on four categories, i.e., angry, sad, happy, and others, using Twitter data. Malte and Ratadiya [21] constructed the BERT model on three classes (overtly aggressive, covertly aggressive, and non-aggressive) to detect cyber abuse in English and Hindi texts, which received an F1 score of 0.6244 and 0.6596, respectively. Using BERT-large on three datasets: ISEAR, SemEval, and Emobank, Park et al. [26] cat- egorized emotions categorically and predicted the valence-arousal dominance (VAD) scores of dimensional emotions. For categorically identifying texts in the ISEAR and SemEval data, their model received micro F1 values of 0.688 and 0.695, respectively, and VAD scores of 0.659, 0.327, and 0.287. To identify emotions and propaganda, Vlad et al. [34] presented an ensemble of BERT, bi-LSTM, and capsule models that received a micro F1 score of 0.5868. Kazameini et al. [18] used BERT word embeddings with bagged SVM to predict author personality attributes. They were 59.03 percent accurate on average.

To the best of our knowledge, there is only one earlier work on Bangla-based transformer models. Das et al. [1] assessed three transformer models: m-BERT, Bangla-BERT, and XLM-R on their own developed Bangla corpus BEmoC, along with several ML and DL models. Among all the models in their work, they got the highest weighted F1 score of 69.73%. Tripto et al.

[33] performed emotion detection in six classes in their custom-made dataset containing 2890 comments from different types of YouTube videos in Bangla, English, and Romanized Bangla. Using DL models such as LSTM and CNN and ML models such as NB and SVM, the best model gives a weighted F1 score of 53.54%.

## 2.2 Datasets for text-based emotion classification (EC)

There are enormous publicly available large-scale English datasets on various emotion classes from different domains. The EmotionLines dataset [11], consisting of two data subsets (i.e., the Friends and Emotion-Push datasets), was released for the challenge; emotions were classified into four labels, i.e., neutral, joy, sadness, and anger, to classify emotion from texts in English. Each subset includes 1,000 English dialogues. EmoBank [7] is another corpus of 10,000 English sentences balancing multiple genres, annotated with dimensional emotion metadata in the VAD representation format. EmoBank excels with a bi-perspectival and bi-representational design. ISEAR (International Survey on Emotion Antecedents and Reactions) is another widely renowned dataset that contains 7,665 sentences. Over a period of many years during the 1990s, a large group of psychologists all over the world collected data for the ISEAR project [30]. Student respondents, both psychologists and non-psychologists, were asked to report situations in which they had experienced all seven major emotions, i.e., joy, fear, anger, sadness, disgust, shame, and guilt. In each case, the questions covered the way they had appraised the situation and how they reacted. The final data set thus contained reports on seven emotions by close to 3,000 respondents in 37 countries on all five continents. SemEval-2017 Task 4 [29] is a dataset that contains 1,250 texts obtained from Twitter, news headlines, Google News, and other sources that were annotated for Ekman's six basic emotions [13]. Mohammad and Marquez [22] developed a small dataset named WASSA-2017 Emotion Intensities (EmoInt) with four emotion classes (fear, joy, sadness, and anger), which caught the attention of the NLP researchers.

On the other hand, only two publicly available datasets on Bangla emotion classes have been published so far (Table 1). Rahman and Seddiqui [28] presented a manually annotated Bangla emotion corpus named BNEmo, which incorporates a diversity of fine-grained emotion labels such as sadness, happiness, disgust, surprise, fear, and anger. They collected a large amount of raw text data from the user's comments on two different Facebook groups (Ekattor TV and Airport Magistrates) and from the public post of a popular blogger and activist, Sarker. A total of 6,327 comments and reviews were annotated in the six categories. In another recent work by Asif Iqbal et al. [15] named BEmoC, they labelled a total of 7,000 Bangla texts into six basic emotion categories from various sources like Facebook and YouTube comments and posts, online blog posts, Bangla story books, Bangla novels, text conversations, and newspapers. We consider EC to be as important for Bangla dialects as it is for any other dialect. Hence, our focus is to build a BERT-based model that can classify Bangla emotions from texts.

**Table 1** Summary of existing Bangla emotion datasets

| Corpus | Publicly available | Classes | Total data (reviews) |
| --- | --- | --- | --- |
| BEmoC | Yes | 6 | 7000 |
| BNEmo | Yes | 6 | 6327 |
| Bangla Youtube Comments | Yes | 6 | 2890 |

## 3 Methods

### 3.1 Model and architecture

In this section, we present the implementation details of our experimental setup. We fine-tuned m-BERT (Fig. 1), BERT-base-multilingual-uncased [2], as this is one of the well-structured back- bones, especially for low-resource languages, and there is a lack of such robust monolingual Bangla models. It is also the only model that was trained in Bangla up until we performed our experiment. The m-BERT is trained in more than 100 languages and has the largest Wikipedia corpus. Embeddings from BERTbase have 768 hidden units. The BERT configuration model takes a sequence of words or tokens at a maximum length of 512 and produces an encoded rep- resentation of dimensionality 768. When fine-tuned on the downstream sentence classification task, very few changes are applied to the BERTbase configuration. In this architecture, only the [CLS] (classification) token output provided by BERT is used. The [CLS] output is the output of the 12th transformer encoder with a dimensionality of 768. It is given as input to a fully con- nected neural network, and the

---

Softmax activation function is applied to the neural network to classify the given sentence. Thus, BERT learns to predict whether a sentence can be classified as any of the six emotion classes.

[2] https://github.com/google-research/BERT/blob/master/multilingual.md

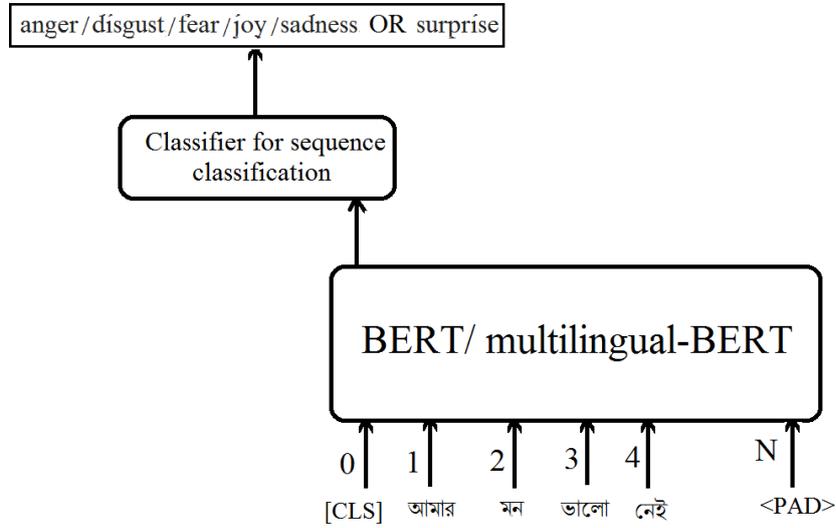

**Figure 1** BERT model for emotion classification in Bangla texts.

In this work, we considered two publicly available Bangla corpora on six major emotion classes, named BNEmo [28] and BemoC [15]. We combined these two datasets with external, manually annotated reviews. We took the reviews from the public YouTube comments on famous Bangla novelist and writer Humayun Ahmed's dramas, "কোথাও কেউ নেই", "□□□□□□□□□", "□□□□□মের রাতে", and comments from BBC Bangla's various Facebook posts on contemporary issues, for instance, war, suicide, hatred, women's abuse, and similar kinds.

In the BNEmo dataset, the six emotion classes were named "sad", "happy", "disgust", "sur- prise", "fear", and "angry". We changed "sad" to "sadness", "happy to "joy," and "angry" to "anger" to keep all the labels similar to the BEmoC dataset.

We performed quite a few operations as part of the preprocessing step. For instance, stop- words are used to remove unnecessary words, fostering programs to deal with crucial ones. Plus, we performed a duplicate drop function if any reviewer mistakenly added to more than one of the above two datasets. After performing such cleaning operations, we finally got 6125 out of 6327 reviews from BNEmo and 6838 out of 7000 reviews from BEmoC (Fig. 2). With our manually annotated reviews, we constructed the Unified Bangla Multi-Class Emotion Corpus (UBMEC), which accumulates a total of 13072 reviews. The data length of UBMEC is 268298 words. The rest of the data is shown in Table 2.

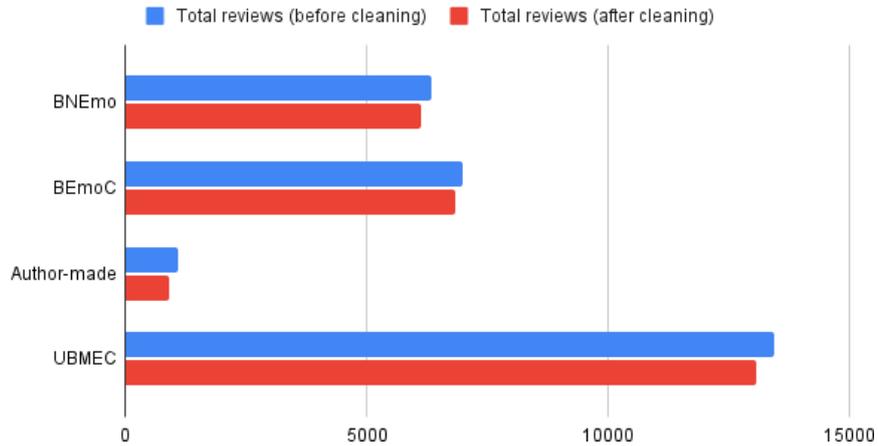

**Figure 2** Total number of reviews from different sources in development of UBMEC.

**Table 2** Description of UBMEC dataset.

| | |
|---|---|
| Data length | 268298 words |
| Data Class | 6 (anger, disgust, fear, joy, sadness, and surprise) |
| Total Reviews | 13072 |
| Size on disk | 1100 KB |
| Train-test data split | 9150, 3922 |
| Maximum words in a single review | 218 |
| Minimum words in a single review | 6 |

The UBMEC dataset consists of 3290 "joy" reviews, 2622 "sadness" reviews, 2422 "anger" reviews, 2049 "disgust" reviews, 1348 "fear" reviews and 1341 "surprise" reviews (Fig. 3).

As mentioned by the authors in [17], emotions can be further categorized into just four classes: joy, sadness, fear/surprise, and anger/disgust. Because the semantic meanings of a sen- tence expressing anger/disgust and fear/surprise are so similar, we divided the dataset into 4 classes to observe the outcomes of the emotion classification using our model. After splitting, we got 3290 "joy" reviews, 2622 "sadness" reviews, 4460 "anger/disgust" reviews, and 2687 "fear/surprise" reviews (Fig 4).

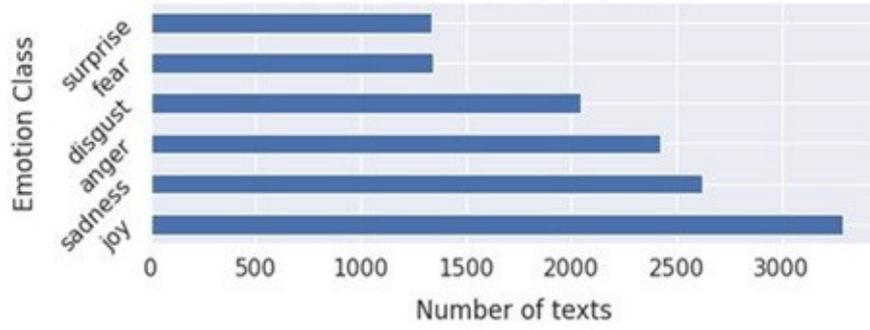

**Figure 3** Distribution of 6 classes in UBMEC Dataset.

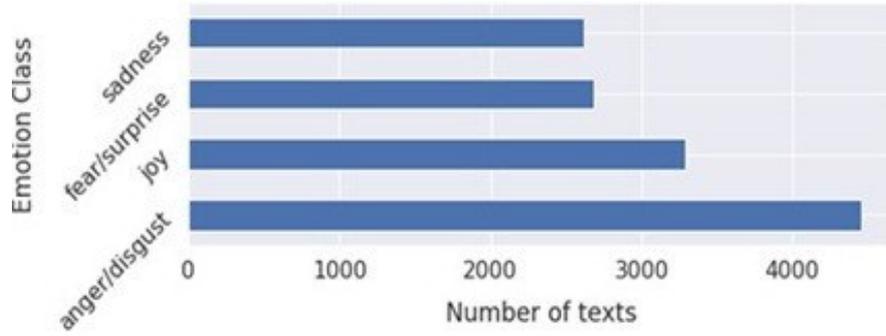

**Figure 4** Distribution of 6 classes in UBMEC Dataset.

From Table 3, we can see sample annotated reviews in the UBMEC dataset for each of the six emotion classes in Bangla sentences.

## 3.2 Model training and fine-tuning

The entire EC is accomplished in two steps: the pre-training of the m-BERT language model took place in the first phase, while the second part involved fine-tuning the outermost clas- sification layer. The Bangla Wikipedia has served as the primary source for pre-training the m-BERT. The m-BERT model was tuned using the training set of the planned UBMEC dataset, which comprises labelled user evaluations. This technique has been used in particular to train the fully linked categorization layer. Language-specific Wikipedia content varies in amount; therefore, data is sampled using an exponentially smoothed weighting (with a factor of 0.7). As a result, low-resource languages are more evenly sampled than high-resource languages like English. Similarly, word counts are sampled to ensure that low-resource languages have an ad- equate vocabulary. Categorical cross-entropy was used as the loss function during training. The lists of hyperparameters used for this research are shown in Table 4.

**Table 3** Sample annotated reviews in UBMEC dataset.

| Emotion Class | Meaning | Text Sample |
| --- | --- | --- |
| Joy | When a writing expresses happy feelings derived from both new and familiar experiences, it is classified as joyful. [25] | এরকম নাটক বারবার দেখেও বিরক্ত লাগে না, এমন লেখক আর দ্বিতীয় একজন হবে না, যার নাম হুমায়ূন আহমেদ |
| Sadness | If a text communicates a response to loss, it falls into the grief category, and feeling sad helps us convince others that we need aid. [25] | সত্যের জয় সবসময় হয় নাহ! এমন হাজারও বাকের আছে মিথ্যের কারণে চলে যেতো হয়! সত্যি কোথাও কেউ নেই! |
| Anger | A letter is considered angry if it expresses unjust treatment of us or others. [25] | দেশ যখন মানুষরূপী শুয়োরেরা চালায়। |
| Disgust | When a text conveys a disappointed sensation about anything bad or unpleasant that occurs physically or psychologically, it falls into the Disgust category. [25] | জনগণের ক্রিটিকাল থিংকিং এবিলিটি প্রয়োজনীয়তার তলানিরও নিচে। আই ওয়াশ আর সার্কাজমে এরা বেশি উৎসাহী। |
| Fear | When a literature depicts threats or dangers to our safety or existence, it expresses terror. | দেশে হিন্দু মুসলিমদের দাঙ্গা লাগানোর চেষ্টা চলছে। |
| Surprise | A writing is considered surprising if it describes an unexpected or surprising incident that occurred unexpectedly. [25] | সুইসাইড করাকে জাস্টিফাই করছেন?! |

**Table 4** m-BERT model hyper-parameters.

| Hyper-parameter | Value |
|---|---|
| Learning rate | 2e-5 |
| Batch size | 64 |
| Number of epochs | 5 |
| Adam epsilon | 1e-8 |
| Gradient accumulation steps | 1 |
| Hidden size | 768 |
| Hidden layers | 12 |
| Maximum sequence length | 64 |
| Parameters | 110 M |

## 4 Results and discussions

### 4.1 Result of machine learning models and m-BERT model

We compared the results of the m-BERT model with those of other ML algorithms. First, we see the results of logistic regression (LR), naive Bayes (NB), and XGBoost on datasets BNEmo, BEmoC, and our proposed UBMEC dataset (Fig 5). The train-test split is the same in both the m-BERT model and the ML model. The results of these three datasets using different ML models are given in detail in Table 5. From the table, we get to see that among all ML models, LR gave the best result. In the BEmoC dataset, LR gives 47% accuracy. While the UBMEC dataset exhibits 44% accuracy through the LR method. Among all the ML models, it is evident that the three datasets, along with our proposed one, perform consistently. This also validates the correctness of the formation of the UBMEC dataset.

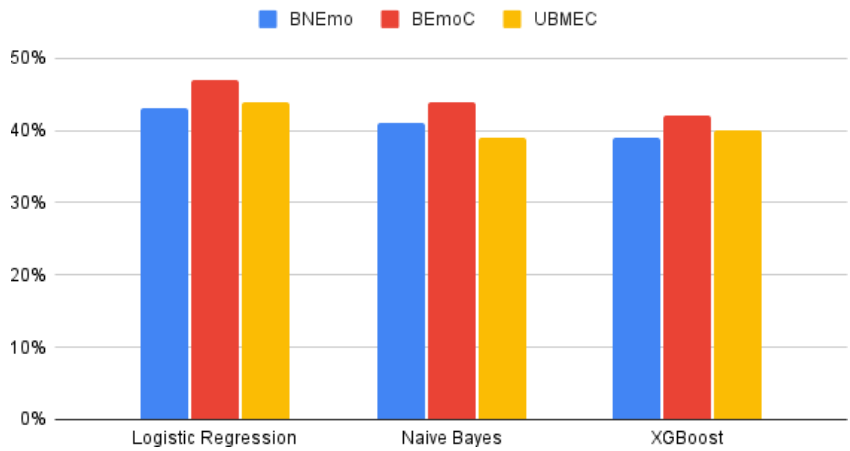

**Figure 5** Accuracy of different machine learning models on three emotion corpus.

ML models have fewer trainable parameters on average than deep neural networks, which explains why they learn so rapidly. Instead of using semantic information, these classifiers use the discriminative power of words with respect to their classes to create class margins. Furthermore, LR outperforms all other ML algorithms since it not only deduces maximum margin hyperplanes but also handles outliers substantially better than other ML algorithms.

On the other hand, DL algorithms are substantially more capable of extracting hidden pat- terns than ML learning classifiers, not simply because they automate the feature engineering process. Quite often, ML algorithms are generally found to be less successful than DL algorithms in this regard.

In view of getting a different insight, along with six classes of EC, we further performed four classes of EC (joy, sadness, fear/surprise, and anger/disgust). For this purpose, we split the UBMEC dataset into train-test data splits of 9,141 and 3,918. This time, the length of the UBMEC dataset is 267847 words. Total reviews (before cleaning, after cleaning) = (13436, 13059). Table 6 shows the results for four classes of EC results in different ML and m-BERT models. The performance metrics indicate the enhancement of results in classifying the four Bangla emotion classes.

**Table 5** Six emotion class results using machine learning and m-BERT models.

| Dataset | Models | Accuracy | Weighted F1-score |
|---|---|---|---|
| BNEmo | LR | 0.41 | 0.45 |
| | NB | 0.40 | 0.47 |
| | XGBoost | 0.38 | 0.44 |
| BEmoC | LR | 0.47 | 0.47 |
| | NB | 0.44 | 0.48 |
| | XGBoost | 0.41 | 0.42 |
| UBMEC | LR | 0.44 | 0.45 |
| | NB | 0.39 | 0.44 |
| | XGBoost | 0.39 | 0.42 |
| | m-BERT | 0.61 | 0.7103 |

**Table 6** Four Emotion classes results using machine learning and m-BERT models.

| Dataset | ML model | Accuracy | Weighted F1-score |
|---|---|---|---|
| UBMEC | LR | 0.53 | 0.55 |
| | NB | 0.51 | 0.56 |
| | XGBoost | 0.49 | 0.53 |
| | m-BERT | 0.69 | 0.76 |

**Table 7** Model performance comparison.

| Methods | Weighted F1 (%) |
|---|---|
| Word2Vec + LSTM [31] | 53.54 |
| XLM-R [35] | 69.73 |
| m-BERT (proposed) | 71.03 |

From Tables 5 and 6, it is observed that the m-BERT model performs well compared to the ML approaches. The highest weighted f1-score for the six Bangla emotion classification classes is 71%, and the highest f1-score for the other four classes is 76%. We limited the training to only seven epochs because the training of m-BERT models requires a substantial amount of time. It is obvious that the model can perform even better if we train it over a larger number of epochs. Additionally, there are other possible approaches. For example, [6] is being used to construct monolingual BERT models dedicated to the Bangla language by enthusiastic Bangla

NLP researchers around the world. As a result, as a potential follow-up to this research, more advanced BERT-based text classification models for Bangla EC can be created, which will surely enrich the field of Bangla NLP research.

### 4.2 Performance comparison with previous models

By assessing the performance in F1-score, we compare the effectiveness of our m-BERT model with the prior techniques [1, 33]. A summary of the comparison is shown in able 7. The findings demonstrate that the proposed m-BERT model performed better than previous approaches for classifying six emotion classes in Bangla by obtaining the highest F1-score (71.03%).

As previously indicated, research on applying BERT techniques to evaluate Bengali emo- tions is limited. In this domain few papers have been published, all of which mostly employed ML or DL classifiers on a small dataset with domain constraints. On the other hand, our dataset comprises more user ratings than previous research and covers numerous genres with six categorization classes: anger, disgust, fear, joy, sadness, and surprise.

## 5 Conclusions

It appears difficult to determine someone's emotional state from their textual communication. Understanding the text's emotions is crucial for human-computer interaction (HCI). In this work, we employed multilingual-BERT (m-BERT) and pre-trained BERT for emotion cat- egorization in Bangla, a language with limited resources. By comparing the outcomes between two earlier, smaller Bangla emotion datasets and our newly created unified dataset using ML and m-BERT classification models, we were able to verify the high quality of our generated dataset. In contrast to traditional word-based ML methods, our analysis demonstrated that using the pre- trained BERT and m-BERT models and optimizing them for downstream Bangla emotion text classification tasks produced higher weighted F1 scores and accuracy metrics.

# Declarations

- Funding: None.
- Conflict of interest/Competing interests: The authors declare that they have no conflicts of interest or competing interests.
- Ethics approval: Not applicable.
- Consent to participate: Not applicable.
- Consent for publication: Not applicable
- Availability of data and materials: The dataset we constructed in this paper is available at https://github.com/Sakibsourav019/UBMEC-Unified-Bangla-Multi-class-Emotion-Corpus-/blob/main/UBMEC%20CorpusASakib(updated).xlsx
- Code availability: The source code is available at https://github.com/Sakibsourav019/UBMEC-Unified-Bangla-Multi-class-Emotion-Corpus-
- Authors' contributions: M.S.U. Sourav: Conceptualization, Software, Methodology, Writing - original draft; H. Wang: Supervision, Methodology, Writing - review & editing; M.S. Mahmud: Supervision, Methodology, Writing - review & editing; H. Zheng: Validation, Resources.